\documentclass[11pt,a4paper]{article}
\usepackage{amsmath}
\usepackage{hltia}
\usepackage{latexsym}
\usepackage{url}

\usepackage{graphicx}
\usepackage{caption}
\usepackage{subcaption}

\title{An Experimental Study of LSTM Encoder-Decoder Model for Text Simplification\\}
\author{Tong Wang\affilA \and Ping Chen\affilA \and Kevin Amaral\affilA\and Jipeng Qiang\affilB\\
\affilA Department of Computer Science \\
University of Massachusetts Boston \\
\affilB Department of Computer Science, Hefei University of Technology \\
{\tt \{tong.wang001,ping.chen,kevin.amaral001\}@umb.edu}}

\summary{Text simplification (TS) aims to reduce the lexical and structural complexity of a text, while still retaining the semantic meaning. Current automatic TS techniques are limited to either lexical-level applications or manually defining a large amount of rules. Since deep neural networks are powerful models that have achieved excellent performance over many difficult tasks, in this paper, we propose to use the Long Short-Term Memory (LSTM) Encoder-Decoder model for sentence level TS, which makes minimal assumptions about word sequence. We conduct preliminary experiments to find that the model is able to learn operation rules such as reversing, sorting and replacing from sequence pairs, which shows that the model may potentially discover and apply rules such as modifying sentence structure, substituting words, and removing words for TS.   
}

\begin{document}
\maketitle


\section{Introduction}
Text Simplification (TS) aims to simplify the lexical, grammatical, or structural complexity of text while retaining its
semantic meaning. It can help various groups of people, including children, non-native speakers, and people with cognitive disabilities, to understand text better. The field of automatic text simplification has been researched for decades. It is generally divided into three categories: lexical simplification (LS), rule-based, and machine translation (MT)~\cite{wang2016text}. 

LS is mainly used to simplify text by substituting infrequent and difficult words with frequent and easier words. However, challenges exist for the LS approach. First, a great number of transformation rules are required for reasonable coverage; second, different transformation rules should be applied based on the specific context; third, the syntax and semantic meaning of the sentence is hard to retain. Rule-based approaches use hand-crafted rules for lexical and syntactic simplification, for example, substituting difficult words in a predefined vocabulary. However, such approaches need a lot of human-involvement to manually define these rules, and it is impossible to give all possible simplification rules. MT-based approach regards original English and simplified English as two different languages, thus TS is the process to translate ordinary English to simplified English. Neural Machine Translation (NMT) is a newly-proposed deep learning approach and achieves very impressive results~\cite{bahdanau2014neural,cho2014learning,sutskever2014sequence}. Unlike the traditional phrased-based MT system which operates on small components separately, NMT systems attempt to build a large neural network such that every component is tuned based on the training sentence pairs.

NMT models are types of Encoder-Decoder models, which can represent the input sequence as a vector, and then decode that vector into an output sequence. In this paper, we propose to apply Long Short-Term Memory (LSTM)~\cite{hochreiter1997long} Encoder-Decoder on TS task. And we show the LSTM Encoder-Decoder model is able to learn operation rules such as reversing, sorting, and replacing from sequence pairs, which are similar to simplification rules that change sentence structure, substitute words, and remove words. Thus this model is potentially able to learn simplification rules. We conduct experiments to show that the trained model has a high accuracy for reversal, sorting, and sequence replacement. Also, the word embeddings learned from the model are close to its real meaning.

\section{Related Work}
Automatic TS is a complicated natural language processing (NLP) task, it consists of lexical and syntactic simplification levels. Usually, hand-crafted, supervised, and unsupervised methods based on resources like English Wikipedia (EW) and Simple English Wikipedia (SEW)~\cite{coster2011simple} are utilized for extracting simplification rules. It is very easy to mix up the automatic TS task and the automatic summarization task~\cite{wang2016new,rush2015neural}. TS is different from text summarization as the focus of text summarization is to reduce the length and redundant content.

At the lexical level,~\cite{glavavs2015simplifying} proposed an lexical simplification system which only requires a large corpus of regular text to obtain word embeddings to get words similar to the complex word. ~\cite{biran2011putting} proposed an unsupervised method for learning pairs of complex and simpler synonyms and a context aware method for substituting one for the other. At the sentence level,~\cite{zhu2010monolingual} proposed a sentence simplification model by tree transformation based on Statistical Machine Translation (SMT).~\cite{woodsend2011learning} presented a data-driven model based on a quasi-synchronous grammar, a formalism that can naturally capture structural mismatches and complex rewrite operations.

The limitation of aforementioned methods requires syntax parsing or hand-crafted rules to simplify sentences. Compared with traditional machine learning~\cite{di2015prediction,wang2015extended} and data mining techniques~\cite{simovici2015compression,hua2016long,zhuang2016evaluation}, deep learning has shown to produce state-of-the-art results on various difficult tasks, with the help of the development of big data platforms~\cite{wang2014fresh,ren2014cielo}. The RNN Encoder-Decoder is a very popular deep neural network model that performs exceptionally well at the machine translation task~\cite{bahdanau2014neural,sutskever2014sequence,cho2014learning}.~\cite{wang2016text} proposed a preliminary work to use RNN Encoder-Decoder model for text simplification task, which is similar to the proposed model in this paper.

\section{The Model}

In this section, we first briefly introduce the basic idea of Recurrent Neural Network (RNN) and Long Short-Term Memory (LSTM), then describe the LSTM Encoder-Decoder model.

\subsection{Recurrent Neural Network and Long Short-Term Memory}
RNN is a class of Neural Network in which internal units may form a directed cycle to
demonstrate the state history of previous inputs. The structure of RNN makes it naturally suited for variable-length inputs such as sentences. For a sequence data $(x_1,...,x_T)$, where at each $t\in \{1,...,T\}$, the hidden state $h_t$ of the RNN is then updated via 

\begin{equation}
h_t = f(h_{t-1}, x_t)
\end{equation}

Where $f$ is the activation function. However, the optimization of basic RNN models is difficult because its gradients vanish over long sequences. LSTM is very good at learning long range dependencies through its internal memory cells. Similar to RNN, LSTM updates its hidden state sequentially, but the updates highly depend on memory cells containing three kind of gates: the forget gate $c_t$ decides how much remembered information to forget, the update gate $i_t$ decides how to update remembered information, the output gate $o_t$ decides how much the remembered information to output.

{\small
\begin{align}
f_t&=\sigma (W_f (x_t, h_{t-1})) \\
i_t&=\sigma (W_i (x_t, h_{t-1})) \\
c_t^\prime &=\tanh (W_c(x_t, h_{t-1})) \\
c_t & = i_t\odot c_t^\prime + f_t\odot c_{t-1} \\
o_t &=\sigma (W_o(x_t, h_{t-1})) \\
h_t &=o_t\odot \tanh(c_t)
\end{align}
}

Recent works have proposed many modified LSTM models such as the gated recurrent unit (GRU)~\cite{cho2014learning}. However,~\cite{greff2015lstm} showed that none of the LSTM variants can improve upon the standard architecture significantly. In this paper, we use the standard LSTM structure in our model.

\graphicspath{ {} }
\begin{figure*}[ht]
\begin{center}
\includegraphics[width=0.7\textwidth,height=0.38\textwidth]{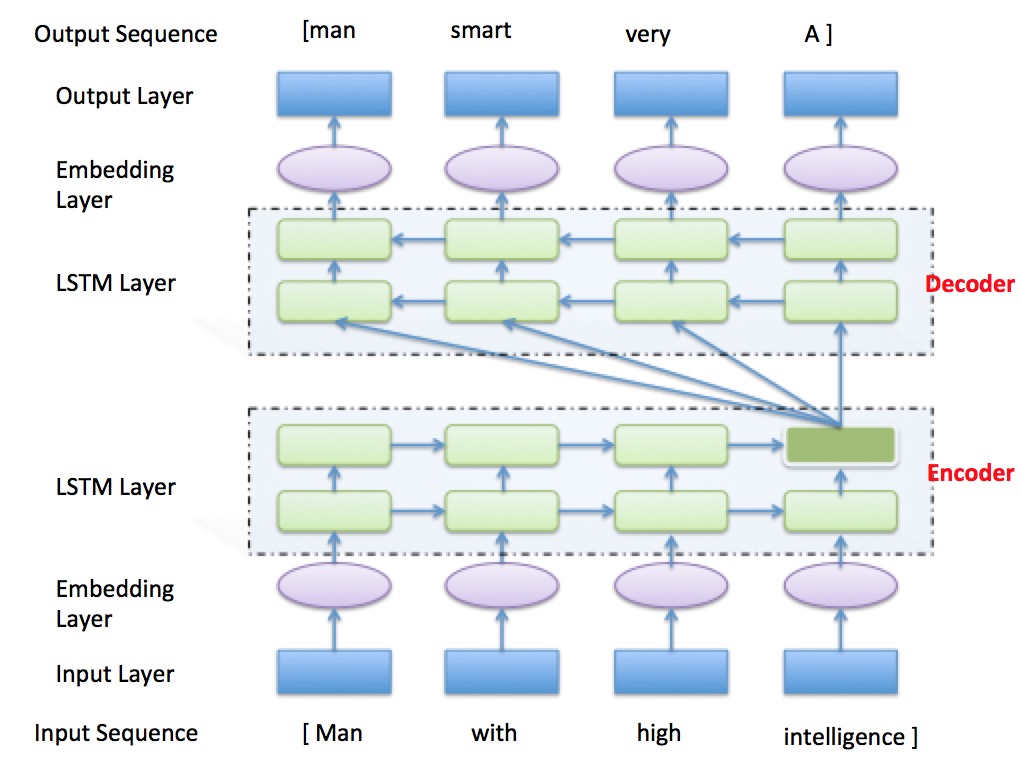}
\caption{LSTM Encoder-Decoder Model}
\label{img:lstm}
\end{center}
\end{figure*}

\subsection{LSTM Encoder-Decoder Model}

Given a source sentence $X = (x_1, x_2, ...,x_l)$ and the target (simplified) sentence $Y = (y_1, y_2, ..., y_{l^\prime})$, where $x_i$ and $y_i$ are in the same vocabulary, $l$ and $l^\prime$ are the length of each sentence. Our goal is to build a neural network to model the conditional probability $p(Y|X)$, then train the model to maximize the probability.

We show our LSTM Encoder-Decoder model in Figure~\ref{img:lstm}. This model uses one-hot representation of words in the sequence in the input layer, and converts it to a 300-dimensional vector in the following embedding layer. We find  that adding an embedding layer can significantly improve performance when the vocabulary becomes large. Then we feed word embeddings through two LSTM layers, and get a vector representation of the input sequence after finishing reading all the words. Finally, we decode this vector to output sequence through two LSTM layers and one output embedding layer.

Let us take the input sentence ``Man with high intelligence" as a difficult sentence, the output sentence ``a very smart man" as the simplified sentence, and represent the pair of sentences as a pair of word indices (we made up some indices here), we have:
\begin{itemize}
\setlength\itemsep{0em}
{\scriptsize
\item[][Man,with,high,intelligence]$\Rightarrow$[A,very,smart,man]
\item[]$[15,27,6,18]\Rightarrow[1, 2, 12, 15]$
}
\end{itemize} 
We only apply sorting, reversing, and replacement to the indices to simplify a sentence, where sorting and reversing could be highly related to changing the structure of a sentence or simplifying a grammar, replacement could be highly related to lexical simplification or removing redundant words. Motivated by this observation, we conduct experiments to show the LSTM Encoder-Decoder is able to learn these three rules automatically, and thus can potentially perform text simplification.

\section{Experiments}

In this section, we conduct experiments to show LSTM Encoder-Decoder can perform basic operations for sequence data. Intuitively, TS should include operations like replacing difficult words with easier words, removing redundant words, simplifying syntax structure by changing word order, etc. In the following experiments, we show that a very basic LSTM Encoder-Decoder model is able to reverse, sort, and replace the elements of a sequence.

We implement the LSTM Encoder-Decoder in Keras~\cite{chollet2015}. The model contains two LSTM layers for both the encoder and the decoder, the output is fed into a softmax layer. RMSprop~\cite{tieleman2012lecture}, which generates
its parameter updates using a momentum on the rescaled gradient, was used as the optimizer in out experiment since it achieves the best performance compared to other optimization methods. We utilized early stopping with patience 5 to avoid over-fitting.

We generate sequences of random integer numbers with length 25 as inputs, since sentences are usually less than 25 words. These integers are the indices of words in the vocabulary $V$. We use three different vocabularies in our experiment $[0,\dots,9],[0,\dots,99],[0,\dots,999]$. For the target outputs, we reverse, sort, and replace words in the input sequence to simulate changing the sentence structure, replacing words, and removing words. The results show that the LSTM Encoder-Decoder is able to learn the reversing, sorting, and replacement operation rules from the provided data, and thus has the potential to simplify a complex text. We use a short example below. 

\begin{itemize}
\setlength\itemsep{0em}
{\scriptsize
\item[] Reverse: $[15,27,6,18,99]\Rightarrow[99,18,6,27,15]$
\item[] Sort: $[15,27,6,18,99]\Rightarrow[6,15,18,27,99]$
\item[] Replace: $[15,27,6,18,99]\Rightarrow[15,7,6,18,19]$
\item[] Combine: $[15,27,6,18,99]\Rightarrow[19,18,15,7,6]$
}
\end{itemize}

\subsection{Reverse}
We first conduct experiments to show that the LSTM Encoder-Decoder can reverse a sequence after training on a large set of sequence pairs $(X, Y)$, where
\begin{align*}
X &= (x_1, x_2, ..., x_{25}) \\ 
Y &= (x_{25}, x_{24}, ..., x_1) \\ 
& x_i\in V
\end{align*}
The results are given in Table ~\ref{table:rev}. $V, H, E$ represent the vocabulary size, the number of hidden neurons in the LSTM layer, and the training epoch, respectively. 

The size of our training data is extremely important for the model. As shown in Table~\ref{table:rev}, the performance decreases significantly if we reduce the size of our training set from 135k to 9k. The size of the vocabulary also influences the performance. A larger vocabulary requires more training data and more hidden neurons in the LSTM layers. By increasing the number of neurons in the LSTM layers from 128 to 256, we produce a higher capacity model with more neurons that can be trained with fewer epochs and achieve a higher accuracy.

In general, this model can reverse an input sequence with higher than $90$ percent accuracy given enough training data. On the other hand, it shows that LSTM is proficient at memorizing long-term dependencies.

\begin{table}
\begin{center}
\caption{Reverse Sequence}\label{table:rev}
{\scriptsize
\begin{tabular}{|c|c|c|c|c|}
\hline
V & H &E & Data &Train,Val,Test\\
\hline
10 &128 &200 &9k,1k,10k &0.9362,0.8732,0.8731 \\
\hline
100 &128 &200 &9k,1k,10k &0.3967,0.1884,0.1883 \\
\hline
100 &128 &200 &135k,15k,10k &0.9690,0.9613,0.9623   \\
\hline
100 &256 &81 &135k,150k,10k &0.9904,0.9784,0.9787  \\
\hline
1000 &256 &133 &135k,15k,10k &0.9410,0.9151,0.9155 \\
\hline
\end{tabular}
}
\end{center}
\end{table}

\begin{table}
\begin{center}
\caption{Sort Sequence}\label{table:sort}
{\scriptsize
\begin{tabular}{|c|c|c|c|c|}
\hline
V & H &E &Train,Val,Test &Train,Val,Test\\
\hline
10 &128 &114 &9k,1k,10k &0.9744,0.9952,0.9956 \\
\hline
100 &128 &200 &9k,1k,10k &0.6370,0.5003,0.4996 \\
\hline
100 &128 &82 &135k,15k,10k &0.9882,0.9907,0.9906  \\
\hline
100 &256 &62 &135k,15k,10k &0.9886,0.9965,0.9969 \\
\hline
1000 &256 &127 &135k,15k,10k &0.9069,0.7958,0.7971 \\
\hline

\end{tabular}
}
\end{center}
\end{table}

\begin{figure*}[ht]
    \centering
    \begin{subfigure}[b]{0.48\textwidth}
        \includegraphics[width=\textwidth, height=0.65\textwidth]{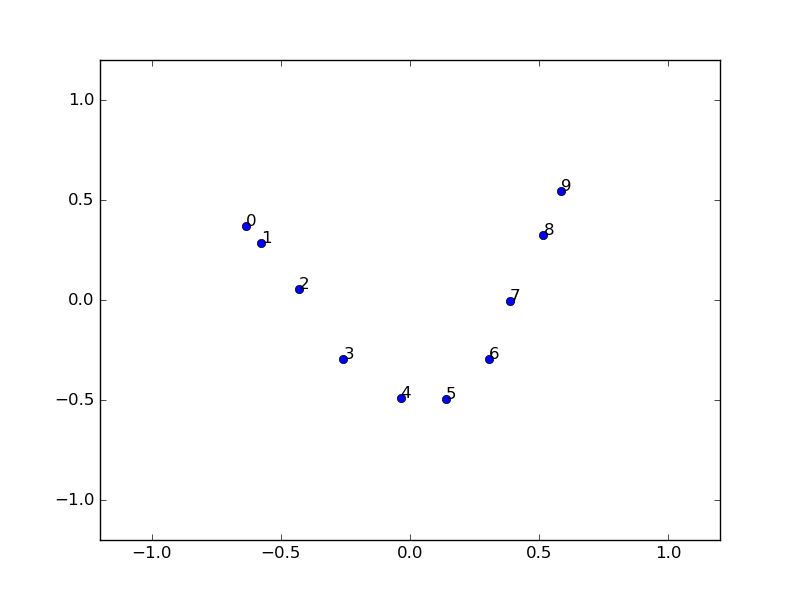}
        \caption{•}
        \label{fig:emb1}
    \end{subfigure}
    ~ 
    \begin{subfigure}[b]{0.48\textwidth}
        \includegraphics[width=\textwidth, height=0.65\textwidth]{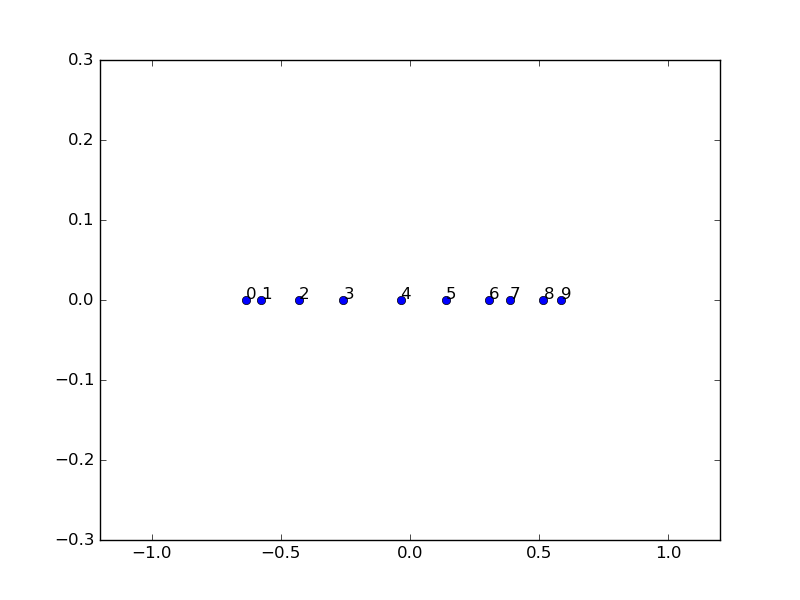}
        \caption{•}
        \label{fig:emb2}
    \end{subfigure}
    ~ 
    \begin{subfigure}[b]{0.49\textwidth}
        \includegraphics[width=\textwidth, height=0.65\textwidth]{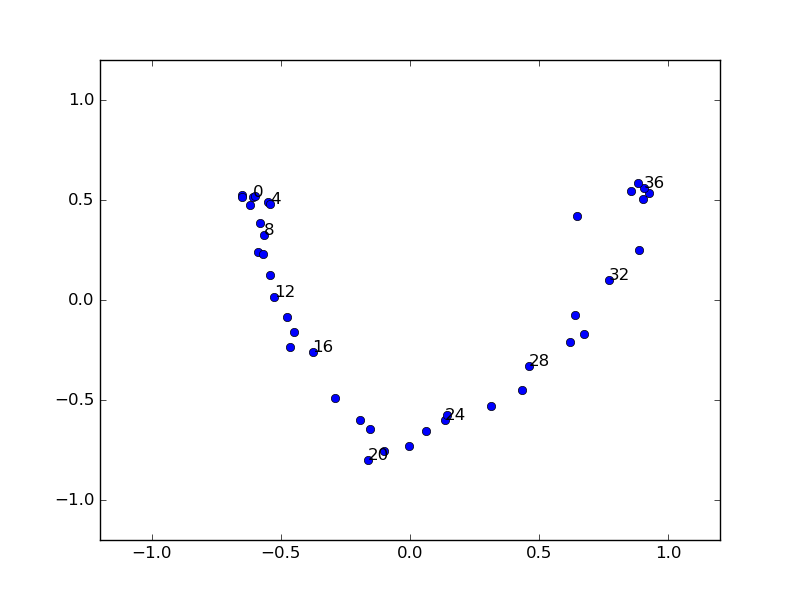}
        \caption{•}
        \label{fig:emb3}
    \end{subfigure}
    \begin{subfigure}[b]{0.49\textwidth}
        \includegraphics[width=\textwidth, height=0.65\textwidth]{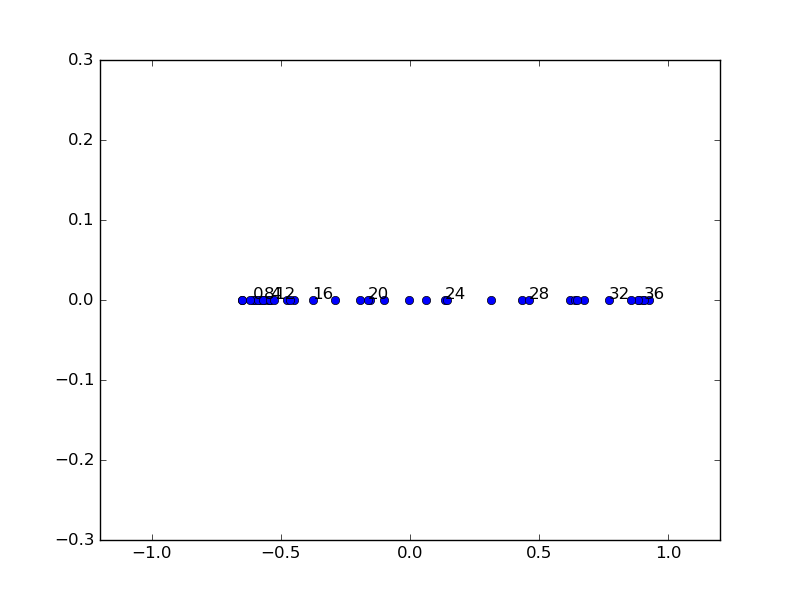}
        \caption{•}
        \label{fig:emb4}
    \end{subfigure}
    \caption{1-dimensional and 2-dimensional PCA projection of number embedding}\label{img:embedding}
\end{figure*}

\subsection{Sort}
Even though Neural Programmer-Interpreters (NPI) ~\cite{reed2015neural}, a recent model, can represent and execute programs such as sorting and addition, the LSTM Encoder-Decoder is much simpler and more light-weight compared to NPI. In the following experiment, we show that the LSTM Encoder-Decoder is able to sort a sequence of integers.

The datasets consist of sequence pairs $(X,Y)$, where 
\begin{align*}
X &= (x_1, x_2, ..., x_{25})\\ 
Y &= \text{sorted}(x_1, x_2, ..., x_{25}) \\
& x_i \in V
\end{align*}
We show the results in Table~\ref{table:sort}. Similarly, the size of the vocabulary, training data, and neurons influence the performance. It is harder to train if we increase the vocabulary to 1000, but the model can still learn the sorting rule with a high accuracy if provided enough training data. 

We extracted the hidden states of the embedding layer from the trained model of the vocabularies of the size 10 and 100. Since the embedding of each word is a 300-dimensional vector, we use Principle Component Analysis (PCA) for dimension reduction and visualize the word embeddings in Figure~\ref{img:embedding}. Interestingly, we find the learned embedding correctly represents the relationship between each word in 1-dimensional and 2-dimensional space. Noted that even though the inputs are integer ``numbers", these numbers are actually symbols, or word indices that should be perpendicular with each other and have same distance. The model does not know, for example, the order between 1 and 2 before training. The model successfully captures the meaning and relationship between words. Similarly, if we provide the model with difficult and simple English sentence pairs, the LSTM Encoder-Decoder may be able to learn how to order words to make the sentence simpler.

\subsection{Replace}
We next show that the LSTM Encoder-Decoder can replace words in a sequence. For the sequence pairs $(X,Y)$, where
\begin{align*}
X &= (x_1, x_2, ..., x_{25}) \\
Y &= (y_1, y_2, ..., y_{25}) \\
y_i & = x_i \bmod n \\
& x_i, y_i \in V
\end{align*}
We let $n = 2, 20, 200$ when $|V| = 10, 100, 1000$. We only keep the top 20 percent of words in the vocabulary, and use these words to replace all matching words in the output sequence. Lexical simplification is similar, in which we can regard the top 20 percent of words in the vocabulary as simple and common words and all the other words are complex words. Similarly, we can also think of the top 20 percent of words as meaningful and important words, other words are redundant. Therefore it also suits the task of removing redundant words. The results are shown in Table~\ref{table:rep}. Compared with the sorting operation, the replacement operation is easier to train when the vocabulary is large, or the size of training data is not large.

\begin{table}
\begin{center}
\caption{Replace in Sequence}\label{table:rep}
{\scriptsize
\begin{tabular}{|c|c|c|c|c|}
\hline
V & H &E & Data &Train,Val,Test\\
\hline
10 &128 &180 &9k,1k,10k &0.9635,0.9172,0.9150 \\
\hline
100 &128 &200 &9k,1k,10k &0.7392,0.5472,0.5488 \\
\hline
100 &128 &61 &135k,15k,10k &0.9927,0.9911,0.9912 \\
\hline
100 &256 &40 &135k,15k,10k &0.9974,0.9997,0.9975  \\
\hline
1000 &256 &75 &135k,15k,10k &0.9868,0.9884,0.9885  \\
\hline

\end{tabular}
}
\end{center}
\end{table}

\begin{table}
\begin{center}
\caption{Combine Three Operations}\label{table:com}
{\scriptsize
\begin{tabular}{|c|c|c|c|c|}
\hline
V & H &E & Data &Train,Val,Test\\
\hline
10 &128 &21 &9k,1k,10k &0.9150,0.8662,0.8660 \\
\hline
100 &128 &130 &9k,1k,10k &0.7570,0.6670,0.6599 \\
\hline
100 &128 &122 &135k,15k,10k &0.9909,0.9985,0.9982 \\
\hline
100 &256 &21 &135k,15k,10k &0.9897,0.9987,0.9988  \\
\hline
1000 &256 &107 &135k,15k,10k &0.9570,0.9405,0.9404  \\
\hline
\end{tabular}
}
\end{center}
\end{table}

\subsection{Combine Three Operations}
We have shown that the LSTM Encoder-Decoder can work well on the reversing, sorting and replacement operations separately, but in reality, a sentence is usually simplified by a complex combination of these three different rules. Therefore, we combine the three operations together to see if this model can still discover the mapping rules between sequences. 

So the data is sequence pairs $(X,Y)$, where $Y$ is obtained by performing modulo for each index in $X$ first, then sorting and reversing. The results are shown in Table~\ref{table:com}. The LSTM Encoder-Decoder continue working very well as expected, and even as good as each of the operations alone. Therefore, the LSTM Encoder-Decoder can easily discover mapping patterns of combined operations between sequences, thus it may potentially find complicated simplification rules.

\section{Conclusion and Future Work}
In conclusion, we find that the LSTM Encoder-Decoder model is able to learn operation rules such as reversing, sorting, and replacing from sequence pairs, which shows the model may potentially apply rules like modifying sentence structure, substituting words, and removing words for text simplification. This is a preliminary experimental study in solving the text simplification problem using deep neural networks. However, unlike the machine translation task, there are very few text simplification training corpora online. So our future work includes collecting complex and simple sentence pairs from online resources such as English Wikipedia and Simple English Wikipedia, and training our model using natural languages.

\bibliographystyle{unsrt}
\bibliography{ref}

\end{document}